\documentclass{article}

\usepackage[preprint]{neurips_2025}

\usepackage[utf8]{inputenc} 
\usepackage[T1]{fontenc}    
\usepackage{hyperref}       
\usepackage{url}            
\usepackage{booktabs}       
\usepackage{amsfonts}       
\usepackage{nicefrac}       
\usepackage{microtype}      
\usepackage{xcolor}         
\usepackage{amsmath}
\usepackage{subfig}
\usepackage{graphicx}
\usepackage{soul}
\usepackage{colortbl} 
\usepackage{array}    

\title{Brownian Bridge Augmented Surrogate Simulation and Injection Planning for Geological CO$_2$ Storage}

\author{%
\textbf{Haoyue Bai}\textsuperscript{1}, 
\textbf{Guodong Chen}\textsuperscript{2}, 
\textbf{Wangyang Ying}\textsuperscript{1}, 
\textbf{Xinyuan Wang}\textsuperscript{1}, 
\textbf{Nanxu Gong}\textsuperscript{1},\\ 
\textbf{Sixun Dong}\textsuperscript{1},
\textbf{Giulia Pedrielli}\textsuperscript{1},
\textbf{Haoyu Wang}\textsuperscript{3},
\textbf{Haifeng Chen}\textsuperscript{3}, 
\textbf{Yanjie Fu}\textsuperscript{1} \\
\textsuperscript{1}Arizona State University, Tempe, USA \\
\textsuperscript{2}Cornell University, Ithaca, USA \\
\textsuperscript{3}NEC Laboratories America, Inc., Princeton, USA \\
\texttt{\{haoyuebai, wying4, xwang735, nanxugong, sixundong,}\\
\texttt{Giulia.Pedrielli, yanjie.fu\}@asu.edu} \\
\texttt{gc594@cornell.edu} \\
\texttt{\{haoyu, haifeng\}@nec-labs.com}
}

\begin{document}

\maketitle

\begin{abstract}
Geological CO$_2$ storage (GCS) involves injecting captured CO$_2$ into deep subsurface formations to support climate goals. 
The effective management of GCS relies on adaptive injection planning to dynamically control injection rates and well pressures to balance both storage safety and efficiency.
Prior literature, including numerical optimization methods and  surrogate-optimization  methods, is limited by real-world GCS requirements of smooth state transitions and goal-directed planning within limited time.
To address these limitations, we propose a Brownian Bridge–augmented framework for surrogate simulation and injection planning in GCS and develop two insights
(i) Brownian bridge as smooth state regularizer for better surrogate simulator; 
(ii) Brownian bridge as  goal-time-conditioned planning guidance for better injection planning.
Our method has three stages: 
(i) learning deep Brownian bridge representations with  contrastive and reconstructive losses from historical reservoir and utility trajectories,
(ii) incorporating Brownian bridge-based next state interpolation for simulator regularization 
(iii) guiding injection planning with Brownian utility-conditioned trajectories to generate high-quality injection plans.
Experimental results across multiple datasets collected from diverse GCS settings demonstrate that our framework consistently improves simulation fidelity and planning effectiveness while maintaining low computational overhead.

\end{abstract}

\section{Introduction}
Geological CO$_2$ storage (GCS) is a technique that involves injecting captured carbon dioxide into deep subsurface formations, such as saline aquifers and depleted reservoirs, for long-term containment~\citep{bg_3_co2survey,co2_survey2}. 
GCS plays a role in climate change mitigation, energy system resilience, and the transition to carbon neutrality. 
GCS operates by capturing, compressing, and injecting CO$_2$ into the deep subsurface through a sequence of injection events while controlling injection rates and well pressures over time. 
Effective GCS management needs to ensure: i) safety: preventing excessive reservoir pressures, leakage, and induced seismicity, and ii) efficiency: optimizing the reservoir’s storage capacity and improving injectivity. 
One of the essential tasks in GCS is adaptive injection planning that dynamically controls CO$_2$ injection rates and well pressures over time to ensure safety and efficiency.


Solving the adaptive injection planning requires adaptive simulation of evolving reservoir states over time and optimization of the injection plan~\citep{adaptive}. 
In prior literature, numerical simulation and optimization based methods can provide accurate results but are computationally intensive and prohibitively slow for real-time or large-scale applications~\citep{num_sim_opt1,num_sim_opt2,nsim_high_cost_1,nsim_high_cost_2}.  
To alleviate these issues, integrated surrogate modeling with optimization algorithms has emerged as an effective alternative~\citep{surrogate_opt1,surrogate_opt2}.  
Surrogate models can provide fast and accurate approximations of high-fidelity simulations, reduce computational costs, and enable iterative optimization in complex storage scenarios.

Real world GCS practices impose two new requirements of adaptive injection planning for deployments: 
i) In physical systems like CO$_2$ injection and geochemical reactions in GCS, state transitions are typically smooth and gradual; Without mechanisms for temporal continuity and smoothing, surrogate models are sensitive to noise and variance in data, often producing abrupt state changes that distort subsurface physics and lead to unreliable predictions; 
ii) GCS planning is inherently goal-oriented and time-constrained: injection planning aims to achieve a target utility (e.g., sufficient CO$_2$ storage, pressure stability) within a limited time.
Thus, injection planning requires the ability to guide decisions toward a target goal within the limited time and to maintain alignment with a consistent, goal-directed trajectory over time.
The two practical requirements call for a new surrogate-optimization framework.

\noindent\textbf{Our Perspective: deep Brownian bridges as smooth state regularizer and goal-time conditioned planning guidance.}  
As a stochastic process, Brownian bridge~\citep{brownia_bridge} provides unique mathematical properties: 
i) \textit{smooth and gradual transitions}: the Brownian bridge models trajectories that are continuous and smooth between a defined start and end point, which aligns with GCS. 
ii) \textit{goal and time-constrained trajectory planning}: the Brownian bridge is conditioned to reach a specific target state at a designated future time and can ensure decision-making to stay on a projected, goal-aligned path.
Along these lines, we identify three insights centered on Brownian bridge to advance surrogate simulation and injection planning: 
i) a deep version of Brownian bridge can learn a Brownian embedding space in which we can incorporate smooth and gradual state transitions into the surrogate simulator, and incorporate goal- and time-conditioned planning into adaptive injection. 
ii) the next state interpolation in Brownian embedding space can serve as an auxiliary supervision signal to regularize the surrogate simulator to learn smooth and physically consistent transitions.
iii) the Brownian bridge projected trajectory can be seen as goal- and time-conditioned guidance in injection planning toward a high storage utility goal yet completed at a specific time. 

\noindent\textbf{Contributions.} 
1) Problem: We tackle the AI for science problem: adaptive injection in GCS as a simulation to optimization task. 
2) Framework: We propose a Brownian Bridge–augmented surrogate simulation and injection planning framework, where the simulator predicts both storage utility and future reservoir states given the current condition and injection plan, while the planner generates injection plans over time.
3) Techniques:  we leverage contrastive and reconstructive losses to learn two deep Brownian bridges, each structured by an encoder, a generator, and a decoder, for reservoir states and storage utilities from observed data;  we leverage  Brownian bridge-interpolated next state as auxiliary supervision to regularize simulator learning; we leverage Brownian bridge-interpolated trajectory as goal-time conditioned guidance for injection planning model learning. 
4) Validations: extensive experiments on multiple CO$_2$ injection datasets demonstrate that our approach consistently achieves superior operational efficiency, improved predictive accuracy, and enhanced strategic robustness compared to state-of-the-art methods.

\section{Preliminaries and Problem Statement}
\begin{figure*}[h]
  \centering
  \subfloat[\centering 
  \label{fig:gcs_overview}
  \small{Geological CO$_2$ Storage}]{
    \includegraphics[width=0.23\textwidth]{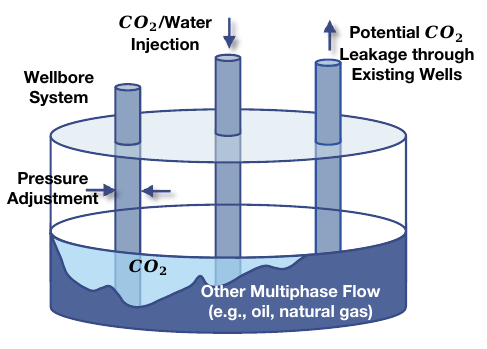}
  }
    \subfloat[\centering 
    \label{fig:lifecycle}
    \small{Reservoir State, Injection Plan, Storage Utility}]{
    \includegraphics[width=0.44\textwidth]{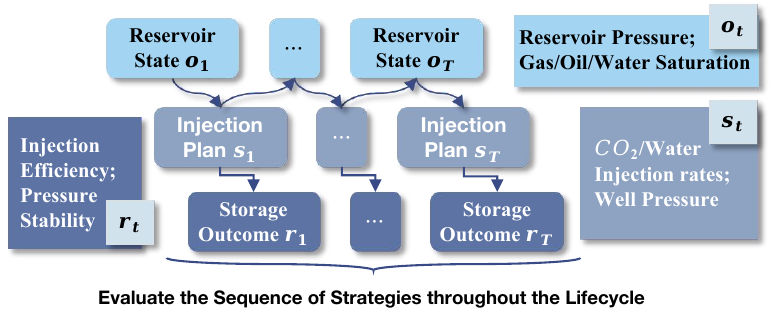}
  }
    \subfloat[\centering 
    \label{fig:BrownianBridge}
    \small{Brownian Bridge}]{
    \includegraphics[width=0.23\textwidth]{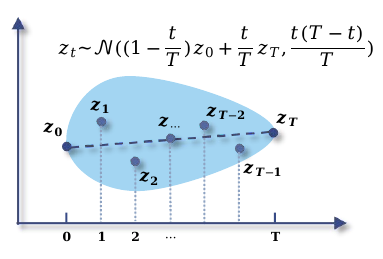}
  }
\caption{\small{Problem and Technique Background}}\label{fig:Preliminaries}
\end{figure*}

\noindent \textbf{Geological CO$_2$ Storage.}
Geological CO$_2$ storage (GCS) is a cornerstone technology in climate change mitigation and energy system decarbonization. 
GCS prevents CO$_2$ release into the atmosphere by injecting it through multiple wells into deep geological formations, such as, saline aquifers and unmineable coal seams (\textbf{Figure~\ref{fig:gcs_overview}}). 
Because geological formations exhibit heterogeneous properties and slow diffusion-driven dynamics, which means the full impact of an injection decision may take months or even years to manifest~\citep{longterm_reason},  effective CO$_2$ sequestration requires multiple injection steps, continuous monitoring, feedback, and adaptive control, often over a long period (e.g., decades).
The entire storage process is structured into a sequence of injection stages, collectively referred to as the \textit{GCS lifecycle}. 
\textbf{Figure~\ref{fig:lifecycle}} shows that each lifecycle consists of multiple discrete time steps. At each step, the system can be characterized by a \textit{reservoir state} and requires the design of an \textit{injection plan}. The utility of this action is observed through a measurable \textit{storage utility}. 

\noindent \textbf{Reservoir State, Injection Plan, and Storage Utility in GCS Systems.}
In GCS, the reservoir state of the storage system includes reservoir pressure, gas saturation, oil saturation, and water saturation at various spatial grid points. These features describe the reservoir’s condition and its response to different operational strategies. 
We define this reservoir state as a vector $\mathbf{o}\in{\mathbb{R}^{|\mathbf{o}|}}$, where $\mathbf{o}_t$ represents the reservoir state at the $t^{th}$ time step.
The system’s operation relies on strategic decisions, such as adjusting CO$_2$ injection rates and well pressure. 
We represent these strategies as a vector $\mathbf{s} \in \mathbb{R}^{|\mathbf{s}|}$, where $\mathbf{s}_t$ is the injection plan at the $t^{th}$ time step.
Each dimension of $\mathbf{s}_t$ is a decision variable of the $t$-th well, for example, the CO$_2$ injection rate of the $t$-th well.
The effectiveness of these strategies is assessed through performance metrics, which capture utilities like injection efficiency, storage integrity, and pressure stability. These metrics form a storage utility vector $\mathbf{r}\in{\mathbb{R}^{|\mathbf{r}|}}$, with $\mathbf{r}_t$ denoting the storage utility at the $t^{th}$ time step. Over a complete lifecycle, these storage utilities provide a comprehensive view of storage performance, revealing whether the strategies achieve long-term stability and effectiveness.

\noindent \textbf{The AI for Science Problem: Surrogate Simulation and Injection Plan Optimization.}
Given a dataset $\mathbb{D} = \{\mathbf{o}, \mathbf{s}, \mathbf{r}\}^{N \times T}$, comprising $N$ GCS trajectories, each with a lifecycle length of $T$ time steps, our task is twofold: 
i) we aim to construct an accurate surrogate simulator $\mathcal{S}$ that maps the reservoir states $\mathbf{o}$ and the strategies $\mathbf{s}$ to corresponding storage utility $\mathbf{r}$. 
ii) we aim to develop a decision-making model $\mathcal{D}$ that generates optimal strategies $\mathbf{s}$ based on the current reservoir state $\mathbf{o}$. The ultimate objective is to optimize the decision model to produce strategies maximizing the average storage utility over the entire lifecycle of the system.

\noindent \textbf{Technical Background: The Brownian Bridge.}  
A Brownian bridge is a stochastic process characterized by Brownian motion conditioned on fixed start and end points~\citep{brownia_bridge} (Figure \ref{fig:BrownianBridge}). 
It models a continuous trajectory that begins at a specified starting value and ends at a designated target value, while evolving stochastically between these two points. 
Formally, given the starting point $z_0$ at the time $t=0$ and the ending point $z_T$ at the time $t=T$, the Brownian bridge at an intermediate time step $t \in [0, T]$ is defined by a Gaussian distribution:
\begin{equation}
z_t \sim \mathcal{N}\left(\left(1 - \frac{t}{T}\right) z_0 + \frac{t}{T} z_T, \frac{t(T - t)}{T}\right).
\end{equation}
This formulation represents a probabilistic interpolation between the initial and final points, with uncertainty maximized midway and minimized at the endpoints.

\section{Proposed Methodology}
\subsection{Framework Overview}
\begin{figure*}[t]
\centering
\includegraphics[width = 1\textwidth]{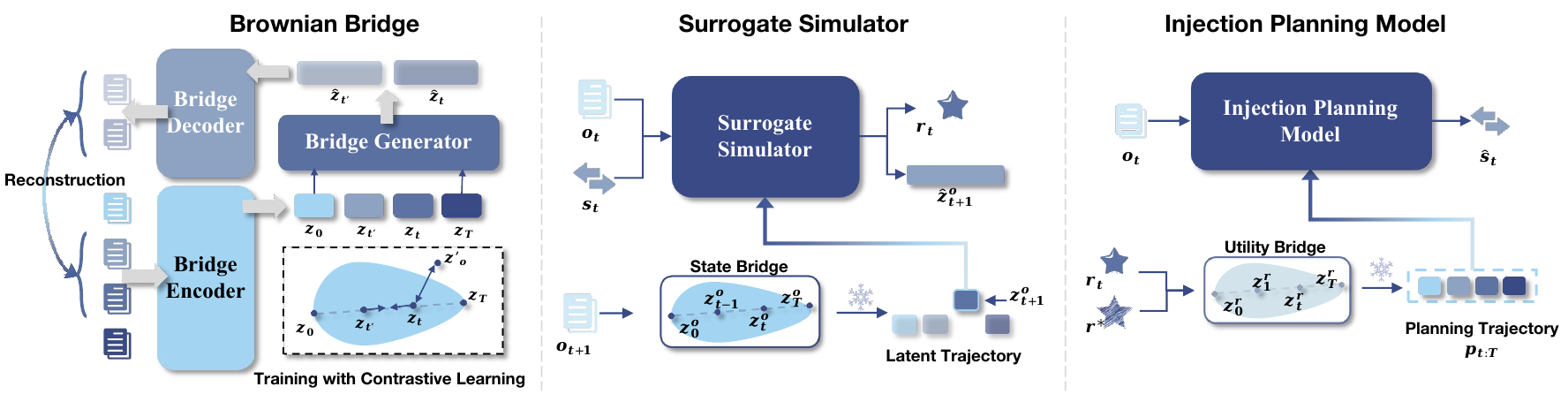}
\caption{\small{Framework}}
\label{fig:framework}
\end{figure*}

To advance GCS management and operations, we propose a Brownian Bridge-enhanced surrogate simulation and injection planning model framework (Figure \ref{fig:framework}), where the surrogate simulator estimates the storage utility and next reservoir state given a reservoir state and an injection plan; the injection planning model generates adaptive injection plans.
We found that Brownian bridge provides two opportunities: i) the ability to enforce smooth state transitions and interpolate the next state can serve as auxiliary supervision and regularization signals to advance next reservoir state estimation; ii) the ability to condition a model to reach a goal can improve knowledge-guided injection planning. 
Our method includes three steps:
\textit{Step 1} integrates contrastive and reconstruction losses to learn two deep Brownian bridges to embed desired reservoir state and storage utility trajectories into a Brownian latent space, so that we can reliably compute desired state and goal-guided utility trajectories. 
\textit{Step 2} incorporates the smooth interpolation of the next reservoir state in Brownian space to regularize surrogate simulator learning. 
\textit{Step 3} leverages the goal-conditioned pursuing ability of Brownian bridges to guide long-term and forward-thinking injection planning. 

\subsection{Deep Brownian Bridge:  GCS Lifecycle State and Utility Dynamics Modeling as Interpolation in Brownian Embedding Space}
Given a start point and an end point, the classic Brownian bridge is a Gaussian process that models the most possible trajectory between two points. 
Inspired by~\citep{bridge_app1,bridge_app2}, we extend the Brownian bridge to a latent representation space, where a trajectory is represented as a sequence of latent embedding vectors, instead of a point sequence in an explicit Gaussian space.  
There are two benefits of using the latent Brownian bridge: 
i) Less susceptible to non-Gaussian scenarios, and more robust and generalized. 
ii) After mapping real system reservoir state or storage utility dynamics to a Brownian space, predicting the next reservoir state or inferring a path toward optimal storage utility is as simple as direct proportional interpolation, instead of using complex parameterized deep recurrent network families. 
In our AI for science problem, given the GCS lifecycle training data of many CO$_2$ injection event sequences, the latent Brownian bridge learning has two tasks:
i) Learn a Brownian encoder to encode the start and end points to two embedding vectors in a latent space; 
ii) learn a Brownian generator to interpolate a latent trajectory between the start and end points. 
We propose to integrate data augmentation, Brownian embedding encoder and interpolation generator, contrastive and reconstruction losses to solve the two tasks.

\noindent\textbf{Data Augmentation to Overcome Science Data Scarcity.}  
We need large training data, for example, the injection event sequences (e.g., injection plan, state changes, utility) of many GCS lifecycles, to train a Brownian bridge to understand, represent, and infer the reservoir state and storage utility dynamics of a GCS system.  
However, GCS lifecycle data are scarce because i) real-world earth science data are limited and ii)  numerical simulation is costly~\citep{nsim_high_cost_1,nsim_high_cost_2}. 
With limited GCS data available,  we develop a two-step data augmentation method to generate more training data from existing GCS lifecycle event sequences and enable self-supervised learning to represent and reason Brownian bridges. 
The first step is sampling: given a GCS lifecycle (i.e., a complete state-injection-utility injection event sequence), we randomly sample diverse fixed-length subsequences. 
The second step is diversification: we add controlled Gaussian noises to each sampled subsequence to create more diversified subsequences to enrich training data. Here, we can control how much noise to add (stochastic variation) by a scaling factor $\alpha$.  

\noindent\textbf{Learning Reservoir State and Storage Utility Related Deep Brownian Bridges.}  
We propose to learn a deep encoder-generator-decoder model to map a sequence of dynamic reservoir state vectors or storage utility vectors into a latent embedding space, where the path between two points is defined and regularized by the Brownian bridge. 

\noindent\underline{\it State Bridge: Brownian Encoder, Generator and Decoder for Reservoir State Sequences.} 
We aim to construct a state-related deep Brownian bridge for reservoir state sequences. The model includes three parts: 
i) The state-related Brownian encoder is to  map an original state vector at a time step into a latent embedding, given by: 
$\mathbf{z}_{t} = \mathcal{B}_{e}^{o}(\mathbf{o}_t)$, where $\mathcal{B}_{e}^{o}$  is the Brownian encoder, $\mathbf{o}_t$ is the reservoir state at time step $t$, $\mathbf{z}_{t}$ is the latent embedding. 
ii) the state-related Brownian generator is to interpolate a smooth trajectory between the embedding of the start point and  the embedding of the end point under the definition of Brownian bridge, given by: 
\small{
\begin{equation}
\hat{\mathbf{z}}_{t} = \mathcal{B}_{g}^{o}(\mathbf{z}_{0}, \mathbf{z}_{T}, t) = \left(1 - \frac{t}{T}\right)\mathbf{z}_{0} + \frac{t}{T} \mathbf{z}_{T},
\end{equation}}
where $\mathcal{B}_{g}^{o}$ is the Brownian generator 
and $\hat{\mathbf{z}}_{t}$ is generated latent embedding at time $t$; 
iii) the state-related Brownian decoder is to reconstruct original reservoir state vectors over time from embeddings over time, given by: $\hat{\mathbf{o}_t}=\mathcal{B}_{d}^{o}(\hat{\mathbf{z}}_{t})$, where $\mathcal{B}_{d}^{o}$ is the Brownian decoder and $\hat{\mathbf{o}}_{t}$ is reconstruct original reservoir state vector.  
The objective function includes: i) a reconstruction loss that measures the gap between the Brownian decoder outputs and the original reservoir state input, and ii) a contrastive loss that pulls positive pairs closer in latent space and pushes negative pairs apart, where positive pairs are latent embeddings from the same reservoir state sequence, and negative pairs are sampled from different reservoir state sequences, by minimizing the following:
\begin{equation}
\min \left\| \mathcal{B}_{d}(\hat{\mathbf{z}}_{t}) - \mathbf{o}_t \right\|_2^2 -\log \sum_{i,j,k}\frac{\exp(\text{sim}(\mathbf{z}_i, \mathbf{z}_j)/\tau)}{\exp(\text{sim}(\mathbf{z}_i, \mathbf{z}_j)/\tau) + \sum_{k} \exp(\text{sim}(\mathbf{z}_i, \mathbf{z}_k)/\tau)},
\end{equation}
where  $\mathcal{B}_{d}^{o}$ is the Brownian decoder implemented by a MLP, $\text{sim}(\cdot, \cdot)$ are the cosine similarity between two vectors, and $\tau$ is a temperature hyperparameter that controls the sharpness of the softmax distribution, $\mathbf{z}_i$ is the anchor embedding, $\mathbf{z}_j$ is a positive sample from the same trajectory as $\mathbf{z}_i$, and $\mathbf{z}_k$ are negative samples drawn from other trajectories in the batch.

\noindent\underline{\it Utility Bridge: Brownian Encoder, Generator, and Decoder for Utility Sequences.} 
Similarly, the Brownian bridge model of storage utility sequences, including the utility-related Brownian encoder $\mathcal{B}_{e}^{r}$, generator $\mathcal{B}_{g}^{r}$, and decoder $\mathcal{B}_{d}^{r}$, have the same structure and training method. But the training data is replaced by storage utility vector sequences. 

\subsection{Learning Surrogate Simulator with Smooth Interpolation of Brownian Next Reservoir States as Auxiliary Supervision}
In GCS management and operations, numerical optimization based simulations, like physical solvers, incur substantial computational overhead and are high-cost~\citep{nsim_high_cost_1,nsim_high_cost_2}. 
A data-driven learnable surrogate simulator serves as a low-cost alternative to evaluate the impact (next reservoir state and storage utility) of a CO$_2$ injection plan given the current reservoir state in GCS management.
In GCS, classic numerical simulators estimate the immediate storage utility and next reservoir state, given the reservoir state and injection plan.
Such a simulation method can be enhanced by incorporating an auxiliary task: one-step ahead prediction of the next reservoir state's Brownian embedding, benchmarked by Brownian interpolation. 

\noindent\textbf{Modeling Intuitions.}  Enforcing the simulator to predict the next state in a Brownian latent space, instead of the next state observed in the physical world, can ensure a smooth, gradual state transition path, rather than abrupt jumps. 
This is particularly useful for environments involving physical systems. In GCS, many physical processes (e.g., CO$_2$ injection and subsequent movement, geochemical reactions, and geomechanical effects) involve gradual changes in pressure, saturation, and fluid composition. In contrast, abrupt transitions in state transition would not reflect the actual physics and chemistry occurring in the subsurface, thus leading to unreliable predictions. 
Therefore, we propose two modeling intuitions (i.e., tasks) for learning the surrogate simulator:  i) a good simulator should accurately estimate storage utility given reservoir state and injection plan;  ii) a good simulator should accurately infer the embedding of next reservoir state in the Brownian latent space. 

\noindent\textbf{Simulator Design.}  
To bring in Brownian supervision to guide the surrogate simulator learning, we propose a different design: given the current reservoir state and injection plan, the surrogate simulator should estimate storage utility and the embedding of the next reservoir state in the Brownian latent space. In other words, we introduce a secondary auxiliary objective that requires the simulator to infer the latent embedding of the next reservoir state in a Brownian latent space. 
Formally, let $t$ be the current time, $\mathbf{o}_t$ is the current reservoir state, $\mathbf{s}_t$ is the injection plan, $\hat{\mathbf{r}}_t$ is the predicted storage utility, $\mathbf{o}_{t+1}$ is the next reservoir state observed in GCS, and $\hat{\mathbf{z}}^{o}_{t+1}$ is the Brownian latent embedding of the next reservoir state observed in GCS. The simulator simulates the storage utility and the Brownian latent next reservoir state embedding, given by: 
\begin{equation}
\hat{\mathbf{r}}_t, \hat{\mathbf{z}}^{o}_{t+1} = \mathcal{S}(\mathbf{o}_t, \mathbf{s}_t),
\end{equation}
where $\mathcal{S}$ is a generic estimation function, such as, MLP, graph neural networks, or convolutional neural networks. 

\noindent\textbf{Objective Function.}  
There are two objectives of learning the surrogate simulator: 
1) minimizing the gap between the simulator-estimated storage utility and the real storage utility; 
2) minimizing the gap between the simulator-estimated Brownian embedding of the next reservoir state and the actual latent embedding output by the state-related Brownian encoder of encoding the actual next reservoir state, which is given by: $\mathbf{z}^{o}_{t+1} = \mathcal{B}_{e}^{o}(\mathbf{o}_{t+1})$.
Therefore, the ultimate optimization objective is:
\begin{equation}
\mathcal{L}_{\mathcal{S}} = \left\| \hat{\mathbf{r}}_t - \mathbf{r}_t \right\|_2^2 + \eta \left\| \hat{\mathbf{z}}^{o}_{t+1} - \mathbf{z}^{o}_{t+1} \right\|_2^2,
\end{equation}
where $\left\| \hat{\mathbf{r}}_t - \mathbf{r}_t \right\|_2^2$ is a storage utility estimation loss, $\left\| \hat{\mathbf{z}}^{o}_{t+1} - \mathbf{z}^{o}_{t+1} \right\|_2^2$ is a Brownian next reservoir state embedding inference loss, and  $\eta$ is a hyperparameter balancing two losses.

\noindent\textbf{Solving the Optimization.}  
Reservoir state, injection plan, and storage utility of each time step is used as an instance to provide the necessary input and supervision signal for the surrogate simulator $\mathcal{S}$ and the stochastic gradient descent is performed to update parameters of $\mathcal{S}$ by minimizing  $\mathcal{L}_{\mathcal{S}}$.

\subsection{Learning Brownian Goal Conditioned Injection Plans}
After learning the surrogate simulator, we aim to learn the injection planning model to decide an injection plan including water injection, CO$_2$ injection, and pressure control of injection wells,  at each time step of the GCS lifecycle. Moreover, we want an instant decision not just to optimize instant utility but also future, long-term, total utility with respect to storage efficiency and safety.

\noindent\textbf{Modeling Intuitions.} Deep Brownian Bridge is a neural process that is conditioned to reach a specific final state (e.g., the upper bound of storage utility) at a fixed time, and, moreover, can interpolate the most likely trajectory towards the final state. 
This ability provides two benefits: i) Goal guidance: It can guide the decision model toward high-reward terminal states that are guaranteed to reach a goal (i.e., a high utility) at a target time, which is useful for goal-conditioned planning.
ii) Exploration efficiency:  It can enable more effective transitions of utility-relevant trajectories rather than wasting effort on aimless explorations and drifts.

\noindent\textbf{Incorporating Brownian Goal Guidance into Injection Planning.}  
Classic injection planning models are used to perceive the current reservoir state and project an injection plan, such as decisions on water injection, CO$_2$ injection, and pressure control. 
Unlike the traditional solution, we propose to leverage the utility-related deep Brownian bridge to inform the injection planning model to make decisions by conditioning on not just the current reservoir state but also a forward-looking trajectory about how to pursue a target utility from the current utility. 
The idea is to leverage the utility-related Brownian generator to generate a prospective latent storage utility trajectory guided by the goal of moving from the recent storage utility to the target utility, then integrate both such utility trajectory and current state as inputs of the injection planning model. 
Formally, let  $\mathbf{r}_{t-1}$ be the recent storage utility, $\mathbf{r}^*$ is a predefined storage utility target (e.g., the empirical maximum from historical data or a theoretical optimum), 
the latent utility embedding trajectory in the utility-related Brownian space is given by: $\left\{ \mathcal{B}^{r}_{g}(\mathbf{r}_{t-1}, \mathbf{r}^*, t') \,\middle|\, t' \in [t, T] \right\}$, where $\mathcal{B}^{r}_{g}$ is the utility-related Brownian generator, $t'$ indexes each time step from $t$ to $T$, and $\mathcal{B}^{r}_{g}(\mathbf{r}_{t-1}, \mathbf{r}^*, t')$ is the desired storage utility at the time $t'$. 
The injection planning model produces an injection plan by conditioning on not only the current reservoir state but also the planning trajectory:
\begin{equation}
\hat{\mathbf{s}}_t = \mathcal{D}(\mathbf{o}_t, \left\{ \mathcal{B}^{r}_{g}(\mathbf{r}_{t-1}, \mathbf{r}^*, t') \,\middle|\, t' \in [t, T] \right\}). 
\end{equation}
where $\mathcal{D}$ is the injection planning model, $\hat{\mathbf{s}}_t$ is an injection plan, and $\mathbf{o}_t$ is the current reservoir state.

\noindent\textbf{Objective Function.}  
After the injection planning model generates an injection plan at the $t$-th time step , we exploit the surrogate simulator to estimate the storage utility of the generated injection plan.
Meanwhile, the utility-related Brownian decoder can decode the embedding of the desired utility at the $t$-th timestep in the latent utility embedding trajectory, into a desired storage utility. 
The objective is to learn the injection planning model by minimizing the gap between the estimated storage utility and the desired storage utility over time. 
Formally, let $\mathcal{S}$ be a simulator, $\mathbf{o}_t$ is current reservoir state, $\hat{\mathbf{s}}_t$ is the proposed injection plan at $t$, the estimated utility at the t-th timestep is:
$\mathcal{S}(\mathbf{o}_t, \hat{\mathbf{s}}_t)$.
The injection planning model is optimized by minimizing the gap between the estimated storage utility and the desired storage utility, given by:
\begin{equation}
\mathcal{L}_{\mathcal{D}} = \left\| \mathcal{S}(\mathbf{o}_t, \hat{\mathbf{s}}_t) - \mathcal{B}^{r}_{g}(\mathbf{r}_{t-1}, \mathbf{r}^*, t') \right\|_2^2,
\end{equation}
where  the desired storage utility signal $\mathcal{B}^{r}_{g}(\mathbf{r}_{t-1}, \mathbf{r}^*, t')$  at the $t$-th time step is interpolated from the planning trajectory by the utility Brownian generator. 

\section{Experiment}
We present extensive experimental results on multiple datasets to evaluate the effectiveness of our proposed method. Specifically, we aim to answer the following research questions:\textbf{RQ1}: How well does our method improve the performance of existing surrogate simulation methods? \textbf{RQ2}: How effectively does our method enhance CO$_2$ storage performance compared to baseline approaches? \textbf{RQ3}: How do different technical components contribute to the effectiveness of our method?

\subsection{Experiment Setting}
\noindent\textbf{Datasets.}
The datasets employed in this study were constructed from the high-fidelity numerical simulator ECLIPSE 2016~\citep{eclipse}, focusing on two distinct CO$_2$ storage scenarios that reflect different geological settings and operational complexities.
The first scenario, denoted as \textbf{Homogeneous-WAG} (\textbf{H-WAG}), represents a small-scale, homogeneous sandstone reservoir configured with a five-spot injection-production pattern. 
The geological model is discretized into a Cartesian grid of $60 \times 60 \times 7$ cells, with a uniform cell size and relatively thin layering to approximate near two-dimensional flow behavior. 
The reservoir rock properties were assigned typical values for unconsolidated sandstone formations, with horizontal permeability ranging from several hundred to over one thousand millidarcies, and porosity values predominantly between 0.20 and 0.30. 
An initial reservoir pressure of approximately 2700 psi was established. 
A water-alternating-gas (WAG) injection scheme was employed to control CO$_2$ mobility and improve storage utility. 
The lifecycle in this scenario consists of 120 time steps. We constructed six datasets with increasing scales, containing 50, 100, 150, 250, 350, and 450 full lifecycle trajectories, denoted as \textbf{H-WAG\_1} through \textbf{H-WAG\_6}.
The second scenario, denoted as \textbf{Heterogeneous-Complex} (\textbf{H-COM}), depicts a larger and geologically more heterogeneous system, featuring a $60 \times 60 \times 9$ grid with denser well placement and more complex stratigraphy. 
The rock properties exhibit greater spatial variability, with permeability fields spanning several orders of magnitude and more diverse porosity distributions. 
Initial reservoir pressures were maintained at comparable levels to the H-WAG case. 
The lifecycle in this scenario extends to 240 time steps. 
Four datasets were constructed at different scales, containing 50, 100, 200, and 300 full lifecycle trajectories, and are referred to as \textbf{H-COM\_1} through \textbf{H-COM\_4}. 
The training, validation, and testing sets are divided into 8:1:1 parts according to the number of trajectories.

\noindent\textbf{Evaluation.}
We evaluate both surrogate modeling accuracy and injection plan optimization performance.
\ul{\it For surrogate simulation evaluation}, we measure the accuracy by computing the mean squared error (MSE) between the predicted storage utility and the ground-truth storage utility included in datasets. This provides a direct assessment of how well the surrogate simulator captures the system dynamics across different operational conditions.
\ul{\it For injection plan optimization evaluation}, we define a Storage Performance Index (SPI) based on simulator outputs to assess CO\textsubscript{2} storage performance of a whole lifecycle. 
Specifically, the SPI for each lifecycle is given by: 
\begin{equation}
\small{
\text{SPI} = \overline{\text{FGIR}} - \overline{\text{FGPR}} + \left( (\text{FGIT} - \text{FGPT}) / \text{FGIT} \right) - \sigma_{\text{FPR}}}
\end{equation}
where $\overline{\text{FGIR}}$ and $\overline{\text{FGPR}}$ denote the average CO\textsubscript{2} injection and production rates, respectively, $\text{FGIT}$ and $\text{FGPT}$ represent the cumulative injected and produced CO\textsubscript{2} volumes, and $\sigma_{\text{FPR}}$ is the standard deviation of the reservoir pressure. These parameters can be obtained from the storage utility.
A higher SPI value indicates better CO\textsubscript{2} storage performance, characterized by enhanced injection capacity, minimized leakage, and stabilized reservoir pressure.
We compute the mean SPI across all lifecycles in the test dataset as the final metric. 
We compare the SPI achieved by the injection plan when executed in both the surrogate simulator and the numerical simulator. 
All experiments are conducted on the Ubuntu 22.04.3 LTS operating system, Intel(R) Xeon(R) w9-3475X CPU@ 4800MHz, and 1 way RTX A6000 and 48GB of RAM, with the framework of Python 3.10.4 and PyTorch 2.5.1. For the baseline and our model, we repeat the experiment 5 times and report the average value.

\begin{table}[t]
\centering
\renewcommand{\arraystretch}{1.3} 
\caption{\small{Performance Comparison Across Different Simulation Scenarios}}
\resizebox{\textwidth}{!}{
\begin{tabular}{lcccccccccc}
\hline
Method & H-WAG\_1 & H-WAG\_2 & H-WAG\_3 & H-WAG\_4 & H-WAG\_5 & H-WAG\_6 & H-COM\_1 & H-COM\_2 & H-COM\_3 & H-COM\_4 \\
\hline\hline
CNN & 0.00224 & 0.00272 & 0.00248 & 0.00214 & 0.00163 & 0.00176 & 0.00079 & 0.00073 & 0.00075 & 0.00073 \\
CNN-Ours & 0.00211 & 0.00265 & 0.00235 & 0.00197 & 0.00159 & 0.00140 & 0.00075 & 0.00069 & 0.00071 & 0.00070 \\
\rowcolor{gray!20}
Improvement & 5.87\% & 2.49\% & 5.46\% & 8.20\% & 2.65\% & 20.17\% & 4.99\% & 6.29\% & 4.54\% & 3.95\% \\
\hline
AE-CNN & 0.00230 & 0.00313 & 0.00317 & 0.00246 & 0.00206 & 0.00197 & 0.00134 & 0.00107 & 0.00089 & 0.00081 \\
AE-CNN-Ours & 0.00222 & 0.00290 & 0.00276 & 0.00237 & 0.00189 & 0.00185 & 0.00123 & 0.00096 & 0.00085 & 0.00076 \\
\rowcolor{gray!20}
Improvement & 3.53\% & 7.24\% & 12.97\% & 3.97\% & 8.14\% & 6.38\% & 7.78\% & 9.72\% & 4.18\% & 6.68\% \\
\hline
ConvLSTM & 0.00193 & 0.00259 & 0.00404 & 0.00302 & 0.00243 & 0.00198 & 0.00139 & 0.00081 & 0.00077 & 0.00069 \\
ConvLSTM-Ours & 0.00165 & 0.00230 & 0.00374 & 0.00252 & 0.00188 & 0.00153 & 0.00104 & 0.00072 & 0.00074 & 0.00067 \\
\rowcolor{gray!20}
Improvement & 14.59\% & 11.46\% & 7.64\% & 16.47\% & 22.40\% & 22.74\% & 25.67\% & 10.52\% & 3.77\% & 2.90\% \\
\hline
GNSM & 0.01208 & 0.00688 & 0.00734 & 0.02203 & 0.01550 & 0.01281 & 0.00629 & 0.00375 & 0.00312 & 0.00241 \\
GNSM-Ours & 0.01145 & 0.00559 & 0.00458 & 0.02055 & 0.01450 & 0.01161 & 0.00533 & 0.00318 & 0.00289 & 0.00227 \\
\rowcolor{gray!20}
Improvement & 5.18\% & 18.79\% & 37.62\% & 6.73\% & 6.46\% & 9.38\% & 15.23\% & 15.26\% & 7.40\% & 5.93\% \\
\hline
ANN & 0.01085 & 0.01958 & 0.01610 & 0.00679 & 0.00822 & 0.00543 & 0.03822 & 0.02142 & 0.02064 & 0.01243 \\
ANN-Ours & 0.00947 & 0.01846 & 0.01508 & 0.00611 & 0.00678 & 0.00520 & 0.01416 & 0.00990 & 0.01240 & 0.00576 \\
\rowcolor{gray!20}
Improvement & 12.70\% & 5.71\% & 6.30\% & 10.08\% & 17.46\% & 4.40\% & 62.96\% & 53.78\% & 39.94\% & 53.71\% \\
\hline
\end{tabular}
}
\label{tab:simulation_results_gray}
\end{table}

\noindent\textbf{Baseline Algorithms.}
\ul{\it For comparisons of surrogate simulator methods}, the baseline algorithms include: (1) \textbf{ANN}~\citep{ANN}: a standard feedforward neural network used as a surrogate to approximate CO$_2$ storage responses based on reservoir state and injection plan; (2) \textbf{CNN}~\citep{CNN}: a surrogate model that predicts reservoir saturation distributions using convolutional neural networks; 
(3) \textbf{AE-CNN}~\citep{aecnn}: a fully convolutional encoder-decoder network with dense blocks; 
(4) \textbf{ConvLSTM}~\citep{convlstm}: combining convolutional layers and ConvLSTM cells to model spatiotemporal evolution of reservoir states; 
(5) \textbf{GNSM}~\citep{gnsm}: a graph neural network surrogate model. 
\ul{For comparisons of injection plan optimization methods}, the baseline methods include: 
(1) \textbf{random policy~(RAND)}: generating injection plan randomly and averages the results over 10 repetitions; (2) \textbf{SAC}~\citep{SAC}: a reinforcement learning method that learns a stochastic control policy through environment interactions to optimize long-term rewards; 
(3) \textbf{POMDP}~\citep{POMDP}: a method that optimizes strategies under uncertainty by modeling the CO$_2$ storage process as a partially observable Markov decision process; and 
(4) \textbf{NSGA-II}~\citep{NSGA}: a population-based evolutionary algorithm that performs multi-objective optimization via non-dominated sorting and elite preservation. 
NSGA-II directly optimizes the full injection plan as a fixed-length strategy sequence.
Each candidate is evaluated through full rollout in the surrogate simulator to ensure fair comparison.

\subsection{Effectiveness of Bridge-Enhanced Surrogate Simulation (RQ1)}

This experiment aims to evaluate how well the idea of Brownian bridge interpolated next state as simulator regularization can improve the performance of baseline methods. 
By incorporating this idea into different architectures, we assess whether our method leads to more accurate and temporally consistent predictions over the full CO$_2$ injection lifecycle.
Table~\ref{tab:simulation_results_gray} shows our method consistently improves the performance of all baseline models across different scenarios. 
For instance, in the H-WAG\_6 scenario, the prediction error of the CNN model decreases from 0.00176 to 0.00140 (a 20.17\% improvement), while in the more complex H-COM\_1 scenario, the error of the ANN model is reduced by 62.96\%, from 0.03822 to 0.01416.
These results validate our hypothesis that modeling temporal continuity is critical for accurate surrogate simulation. 
Our method improves performance across various methods and datasets, demonstrating its generalizability and robustness.

\subsection{Effectiveness of Injection Plan Optimization (RQ2)}

\begin{table}[t]
\centering
\renewcommand{\arraystretch}{1.3}
\caption{\small{Comparison of Decision Strategies Across Scenarios}}
\resizebox{\textwidth}{!}{
\begin{tabular}{lcccccccccc}
\toprule
Dataset & H-WAG\_1 & H-WAG\_2 & H-WAG\_3 & H-WAG\_4 & H-WAG\_5 & H-WAG\_6 & H-COM\_1 & H-COM\_2 & H-COM\_3 & H-COM\_4 \\
\midrule
RAND & 0.01224 & -0.87519 & -0.32341 & 0.20419 & -0.18921 & 0.04251 & 0.08970 & -1.05086 & 0.32720 & -0.20124 \\
SAC & 0.74475 & 0.89808 & 0.86930 & 0.88703 & \underline{1.20986} & 1.08951 & 0.88418 & 1.00218 & 0.81055 & \underline{1.28069} \\
POMDP & \underline{0.85544} & \underline{1.07623} & 0.97523 & \underline{1.04814} & 1.15012 & \underline{1.18885} & 0.97278 & \underline{1.29326} & \underline{0.85680} & 1.26131 \\
NSGA-II & 0.72146 & 0.92391 & \underline{1.15187} & 0.92426 & 1.13187 & 0.31259 & \underline{1.20690} & 0.90107 & 0.81491 & 0.97578 \\
\rowcolor{gray!20}
Ours & \textbf{1.35140} & \textbf{1.35223} & \textbf{1.36021} & \textbf{1.36291} & \textbf{1.35927} & \textbf{1.35646} & \textbf{1.33317} & \textbf{1.39324} & \textbf{1.39760} & \textbf{1.39605} \\
\bottomrule
\end{tabular}
}

\label{tab:decision_results}
\end{table}

\begin{figure*}[htbp]
  \centering
    \subfloat[\centering 
    \label{fig:CDF}
    \small{CDF of CO$_2$}]{
    \includegraphics[width=0.2\textwidth]{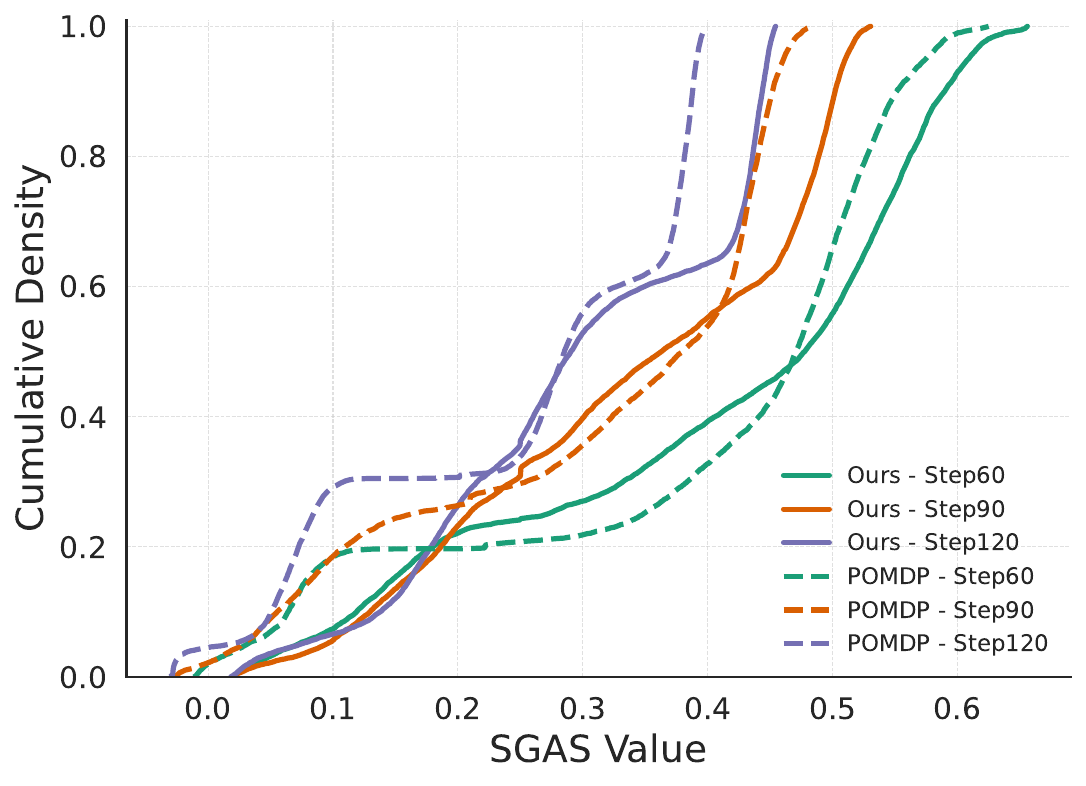}
  }
  \subfloat[\centering 
  \label{fig:Pressure of POMDP}
  \small{Pressure Change of POMDP}]{
    \includegraphics[width=0.19\textwidth]{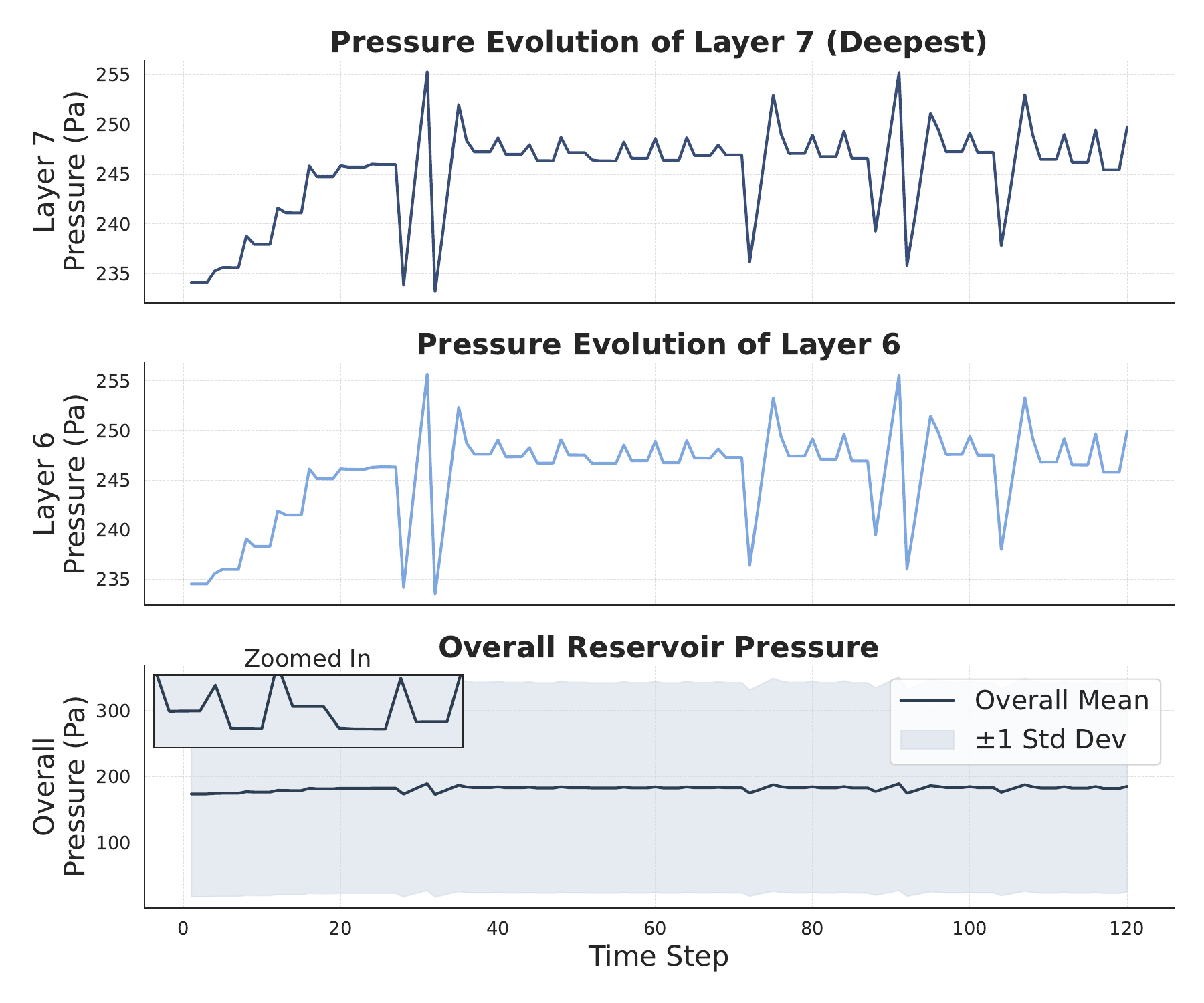}
  }
    \subfloat[\centering 
    \label{fig:Pressure of Ours}
    \small{Pressure Change of Ours}]{
    \includegraphics[width=0.19\textwidth]{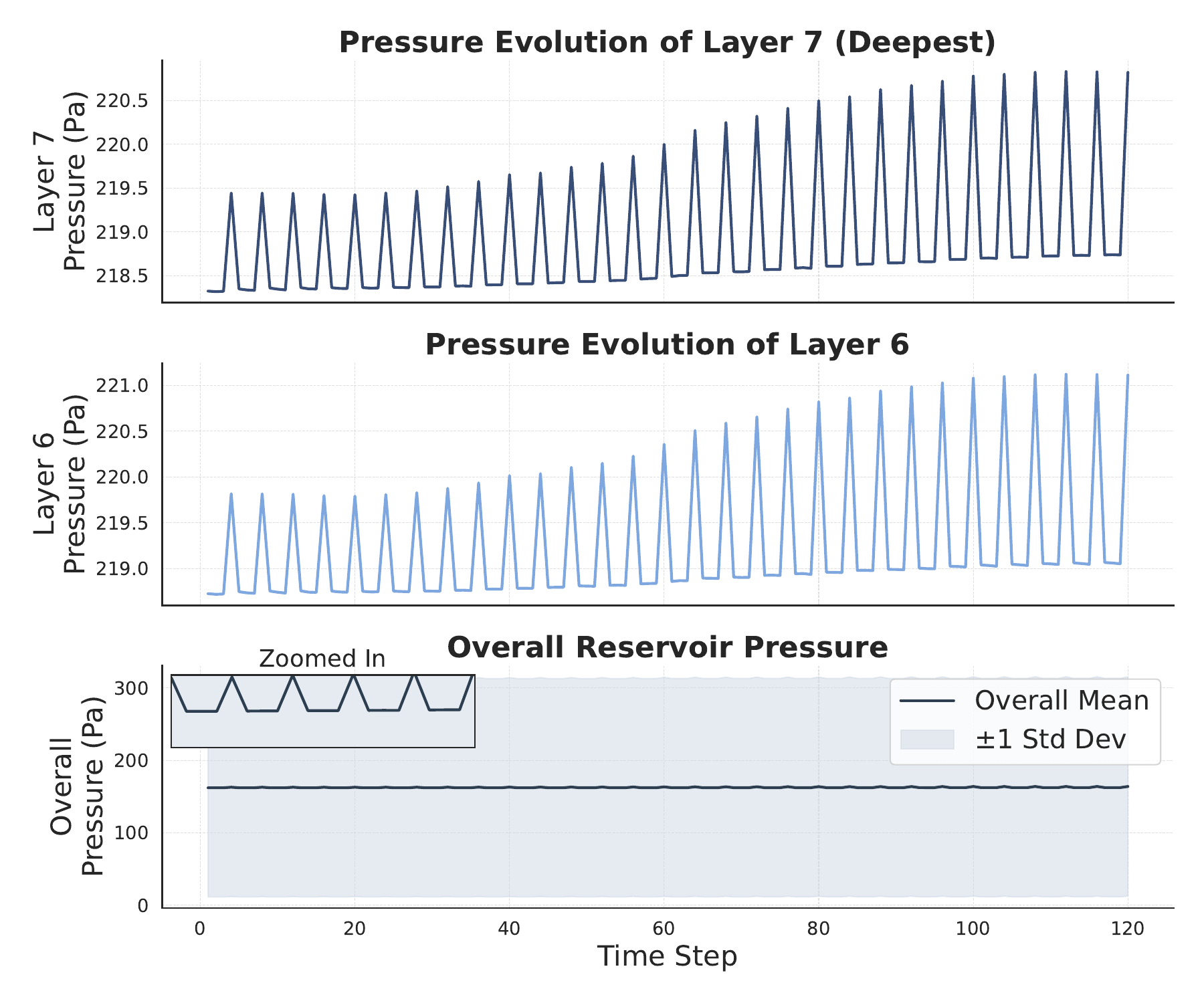}
  }
    \subfloat[\centering 
    \label{fig:3D of POMDP}
    \small{CO$_2$ Distribution of POMDP}]{
    \includegraphics[width=0.19\textwidth]{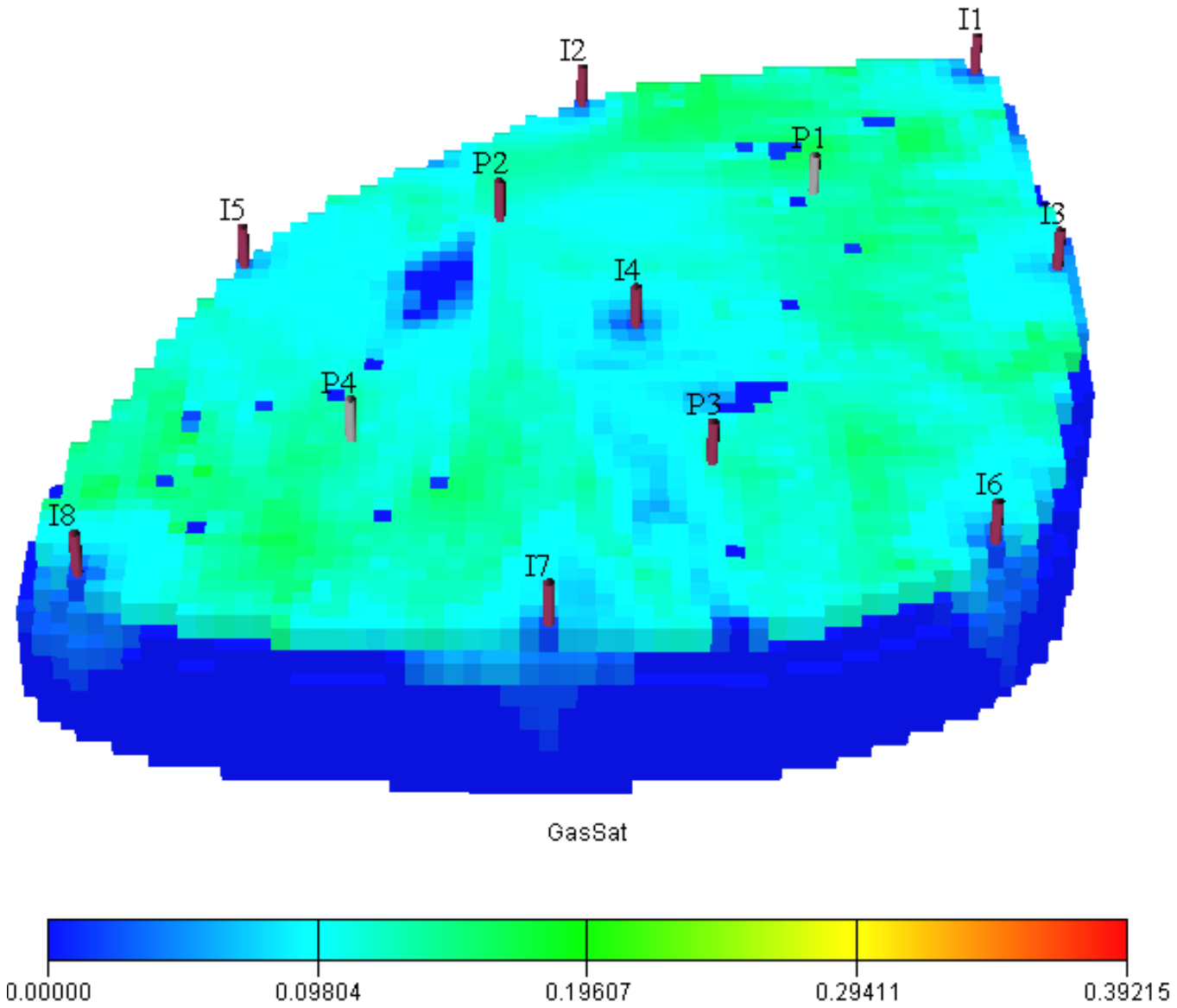}
  }
    \subfloat[\centering 
    \label{fig:3D of ours}
    \small{CO$_2$ Distribution of Ours}]{
    \includegraphics[width=0.19\textwidth]{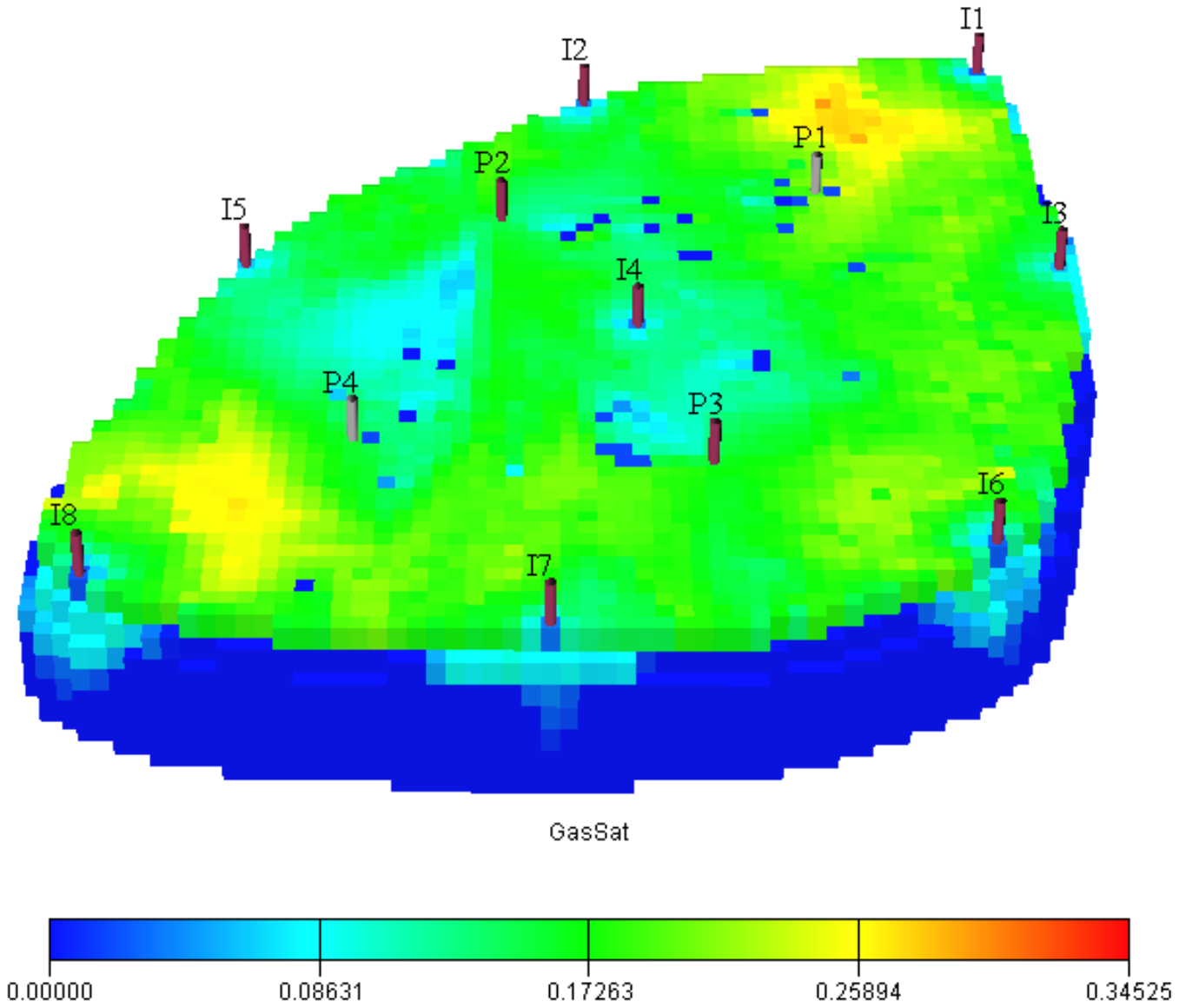}
  }
\caption{\small{Pressure changes and underground CO$_2$ storage distribution}}\label{fig:vis_num_sim}
\end{figure*}

This experiment evaluates the effectiveness of our injection planning model in improving CO$_2$ storage performance. We use the SPI to quantify the injection plan across the whole lifecycle and further validate the results by visualizing the cumulative distribution function (CDF) of CO$_2$, pressure evolution, and spatial CO$_2$ distribution using high-fidelity numerical simulation.
Table~\ref{tab:decision_results} shows our method consistently outperforms all baseline approaches across both scenarios. For instance, in the H-COM\_3 case, our injection plan achieves an SPI of 1.3976, compared to 0.8568 by the strongest baseline (POMDP). 
To further verify the physical plausibility and effectiveness of our method, we selected the strongest baseline (POMDP) and our method to generate multiple injection plans to control the whole lifecycle, which were then evaluated using the ECLIPSE 2016 numerical simulator on the H-COM setting.
Figure~\ref{fig:CDF} shows the CDF of CO$_2$ saturation (SGAS) reveals that our method consistently achieves higher saturation levels across multiple time steps. This indicates that our strategy enables more effective CO$_2$ storage.
Figures~\ref{fig:Pressure of POMDP} and \ref{fig:Pressure of Ours} depict the pressure evolution throughout the lifecycle. Compared to the baseline, our method results in more regular and stable pressure dynamics. This reflects greater control and operational safety, as sudden pressure fluctuations are known to increase the risk of reservoir damage or leakage.
Figures~\ref{fig:3D of POMDP} and \ref{fig:3D of ours} provide a 3D visualization of the underground CO$_2$ distribution at the end of the lifecycle. Due to the lower density of CO$_2$, injected CO$_2$ tends to accumulate in the upper layers. The resulting distribution from our strategy aligns well with this physical expectation, suggesting that our decisions are geophysically consistent. Furthermore, the CO$_2$ plume generated by our method is more uniformly distributed across the reservoir, indicating more balanced injection and less local saturation bias.


\subsection{Component Analysis and Ablation Study (RQ3)}

\begin{figure*}[htbp]
  \centering
  \subfloat[\centering 
  \label{fig:alpha}
  \small{Effect of $\alpha$}]{
    \includegraphics[width=0.2\textwidth]{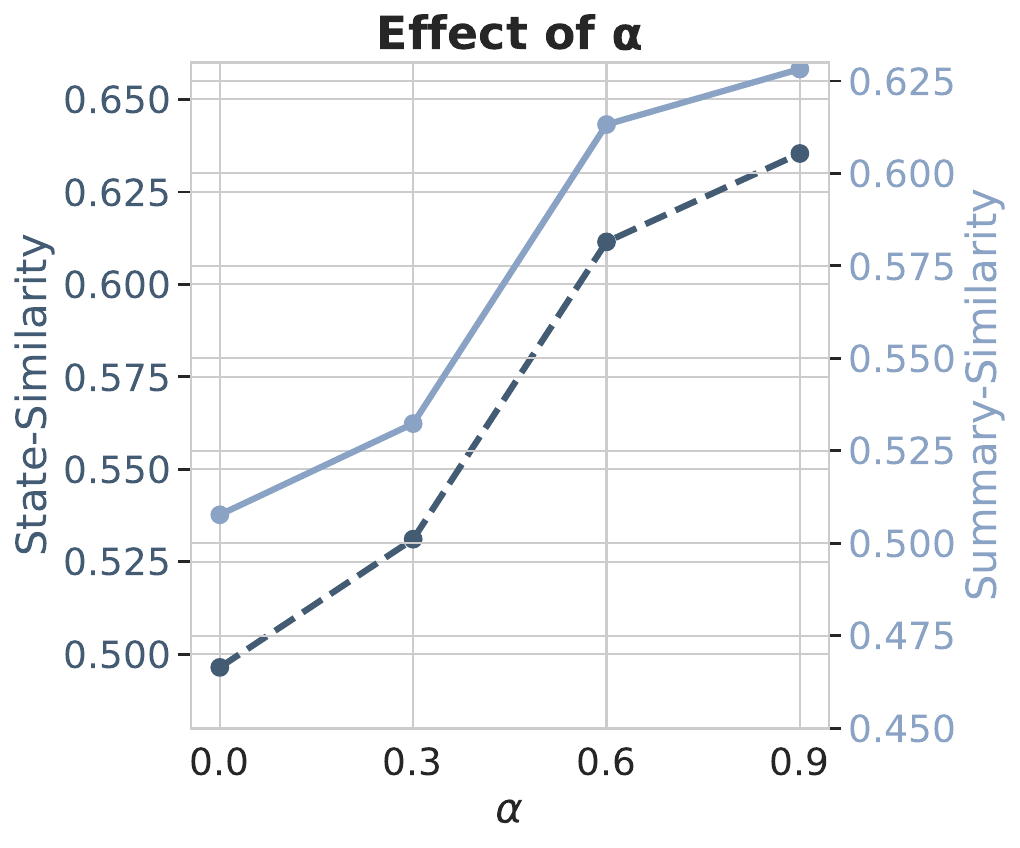}
  }
    \subfloat[\centering 
    \label{fig:eta}
    \small{Effect of $\eta$}]{
    \includegraphics[width=0.19\textwidth]{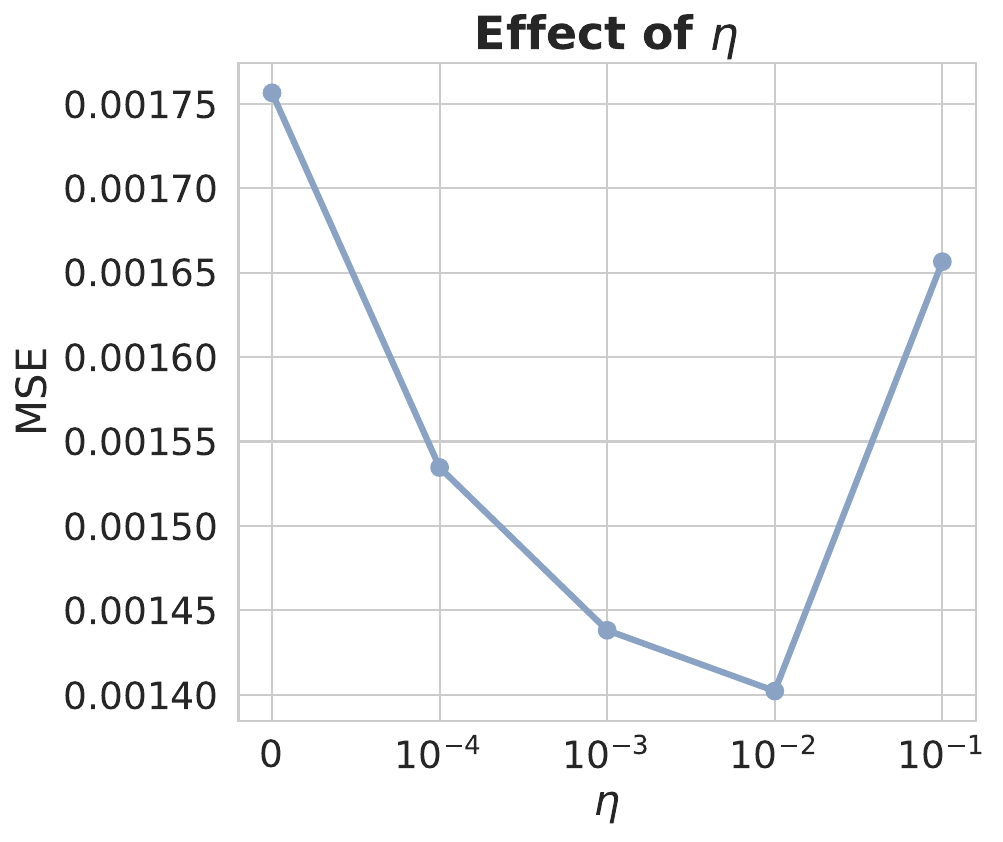}
  }
    \subfloat[\centering 
    \label{fig:abl_wag}
    \small{Decision Ablation}]{
    \includegraphics[width=0.19\textwidth]{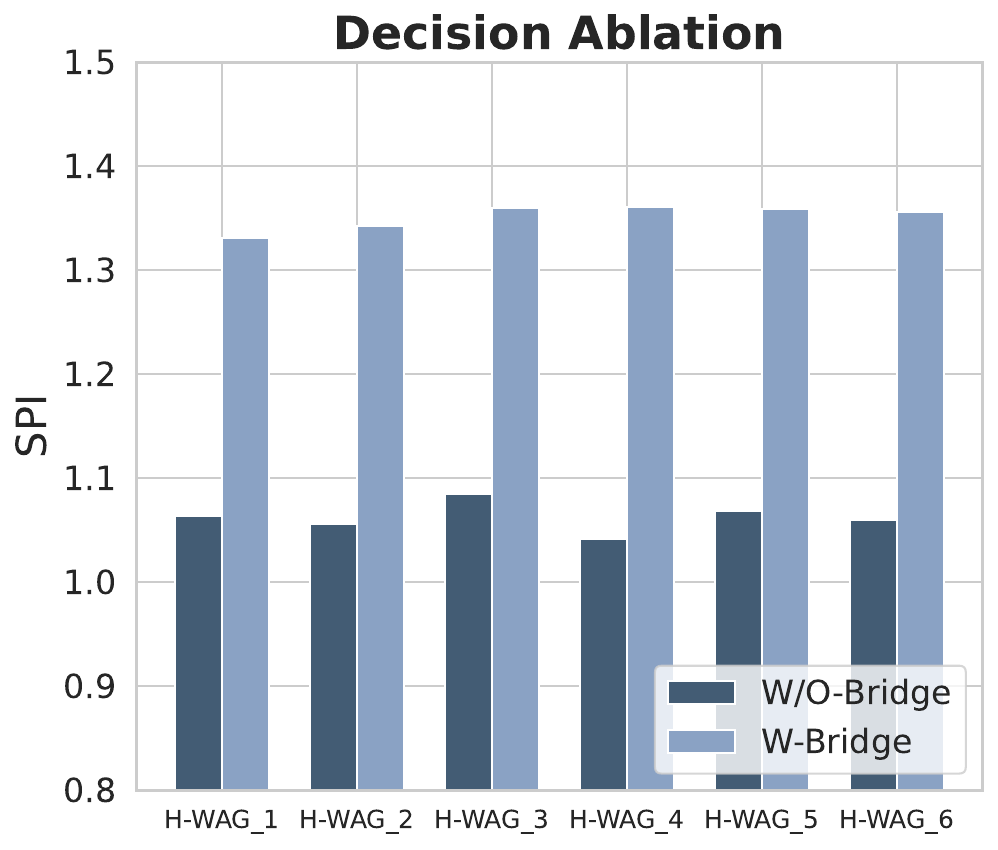}
  }
  \subfloat[\centering 
    \label{fig:times}
    \small{Simulation Time}]{
    \includegraphics[width=0.19\textwidth]{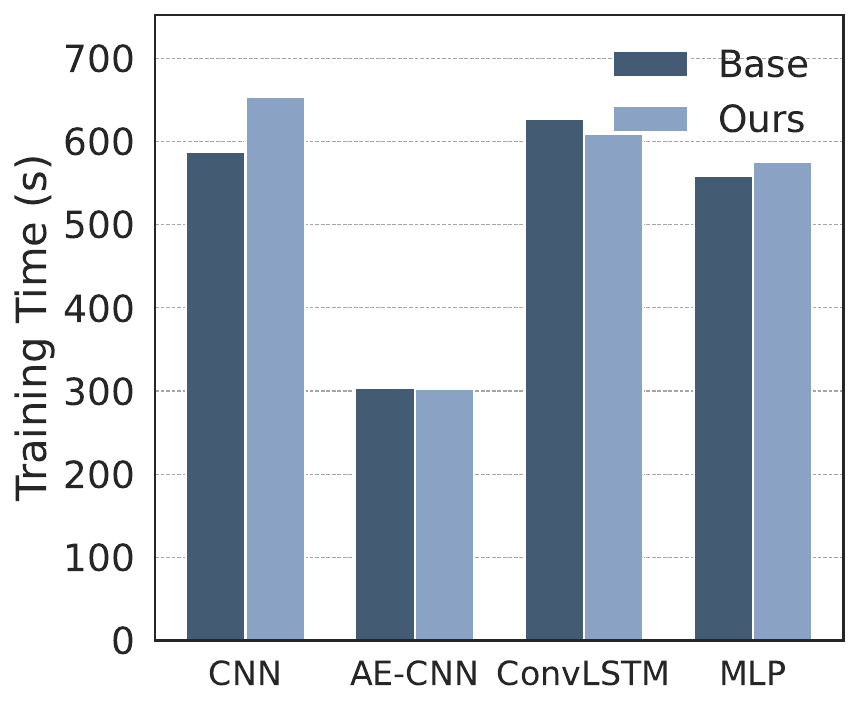}
  }
  \subfloat[\centering 
  \label{fig:timed}
  \small{Optimization Time}]{
    \includegraphics[width=0.2\textwidth]{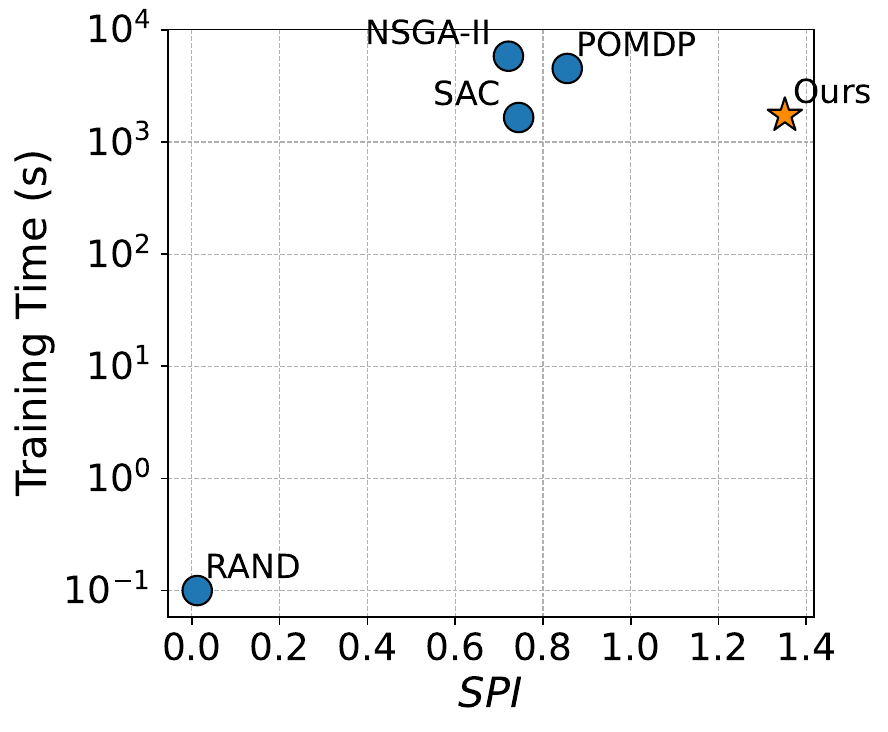}
  }
\caption{\small{Investigation of Proposed Method}}\label{fig:Investigation}
\end{figure*}

Figures~\ref{fig:alpha} show the effect of trajectory construction parameters $\alpha$, which control the noise strength. We evaluate the quality of the generated trajectories by computing their cosine similarity to ground-truth reservoir state and storage outcome trajectories. Results indicate that moderate to high values of $\alpha$ yield higher similarity scores.
Figure~\ref{fig:eta} examines the sensitivity of surrogate simulation to the auxiliary task weight $\eta$ on the H-WAG\_6 case. We observe a U-shaped performance curve, with the lowest prediction error occurring near $\eta = 10^{-3}$, suggesting that a balanced auxiliary loss contributes to improved learning stability and accuracy.
Figure~\ref{fig:abl_wag} presents the decision-phase ablation results. Across both all H-WAG scenarios, adding bridge-guided trajectory planning significantly improves SPI scores, confirming that trajectory-aware decision guidance is critical for achieving a high-quality injection plan.. 
Figure~\ref{fig:times} reports the training time required to optimize different surrogate models. Ours demonstrates comparable computational efficiency despite the inclusion of an auxiliary task that predicts Brownian-bridge-guided latent representations. This latent space is of significantly lower dimensionality than the original reservoir state, which contributes to improved predictive performance without introducing substantial computational overhead.
Figure~\ref{fig:timed} presents the training time for decision optimization across baseline methods. Our method achieves the highest SPI score while maintaining a relatively low optimization time.

\section{Related Work}
Geological resources play a central role in the sustainable development of global energy systems, involving both the extraction of hydrocarbons~\citep{oil1,SAC} and the geological storage of energy carriers, such as CO$_2$~\citep{bg_3_co2survey,aecnn,co2_survey2}. Effective utilization and management of these geological resources require comprehensive and dynamic modeling approaches that span exploration, production, storage, and abandonment phases ~\citep{Lessons}. Within this broader context, geological CO$_2$ storage (GCS) has gained prominence as a critical technology to mitigate climate change through secure and efficient subsurface containment of greenhouse gases.
Dynamic modeling in GCS encompasses multiphase flow, geomechanical behavior, and geochemical interactions, primarily studied through detailed numerical simulations~\citep{nsimuse,xu2004numerical}. Despite their robust predictive capabilities, these numerical methods demand substantial computational resources, which often limit their practical deployment in operational decision-making scenarios~\citep{nsim_high_cost_1,nsim_high_cost_2}. 
Consequently, researchers have increasingly adopted data-driven surrogate simulator approaches using artificial intelligence and machine learning (AI\&ML) techniques, including Artificial Neural Networks~\citep{ANN}, convolutional neural networks~\citep{CNN,aecnn}, recurrent neural networks~\citep{convlstm}, and graph neural networks~\cite{gnsm}. 
And effective as feature transformation~\citep{ying2024unsupervised,ying2023self,gong2025unsupervised} and feature selection~\citep{ying2024feature,ying2024revolutionizing,wang2024knockoff} techniques have been widely adopted in scientific modeling to improve generalization, reduce noise, and enforce smoothness in surrogate models~\citep{wang2025towards,ying2025survey}.
These intelligent proxy models offer rapid, accurate approximations of numerical simulation outcomes, significantly reducing computational costs~\citep{surrogate_fast1,surrogate_fast2}. Such strategy is also proved to be true in other domains such as traffic~\citep{zhang2019cityflow,da2024cityflower,chen2024syntrac}.
The integration of surrogate models with advanced optimization algorithms, such as Particle Swarm Optimization~\citep{PSO} and Non-dominated Sorting Genetic Algorithm II (NSGA-II)~\citep{NSGA, NSGA2}, has proven effective in optimizing complex geological storage strategies. More recently, reinforcement learning approaches~\citep{SAC,rl2} such as soft actor-critic have been explored for adaptive control of GCS management, enabling agents to learn continuous policies through interactions with surrogate simulators under uncertainty, and offering improved adaptability and long-term performance.
Extending these advancements, our research introduces a novel trajectory-based framework that leverages Brownian bridge methods to generate continuous state and policy trajectories from limited historical data. This approach ensures strategic continuity and improved lifecycle management, supporting robust and forward-looking.

\section{Conclusion Remarks and Limitations}
We present a Brownian Bridge–augmented framework for CO$_2$ injection strategy optimization in geological CO$_2$ storage (GCS), addressing the challenges of temporal continuity and goal-conditioned planning. Our method integrates Brownian bridge representations into both surrogate simulation and injection planning. Specifically, we (i) learn temporally smooth latent representations of reservoir state and utility trajectories using contrastive and reconstructive objectives, (ii) enhance simulation fidelity by interpolating next states via Brownian bridges, and (iii) guide adaptive injection planning through utility-conditioned Brownian trajectories. 
Extensive experiments across diverse GCS datasets validate that our approach improves simulation accuracy and injection plan quality while maintaining low computational overhead. 
A current limitation lies in the multi-stage structure of the framework, which may introduce cumulative errors across modules. Future work will explore tighter integration to mitigate this issue and enhance overall robustness.
Our work attempts to solve real-world problems through a framework that integrates simulation and decision-making. In the future, we plan to expand this thinking to more areas such as basic research~\citep{gong2025neuro,ying2024topology,gong2025evolutionary}, business~\citep{li2023sehf,rs1,wang2024llm,rs2,rs3}, and medicine~\citep{liu2019edta,wang2022successful,liu2024calorie,wang2024lcmdc,liu2024pth,li2024sade}.

\bibliographystyle{plainnat}
\bibliography{reference}


\end{document}